\pdfoutput=1
\documentclass[journal]{IEEEtran}
\usepackage{amsfonts}
\usepackage{amsmath}
\usepackage{amssymb}
\usepackage[numbers,sort&compress]{natbib}
\usepackage[colorlinks, linkcolor=red, anchorcolor=blue, citecolor=green]{hyperref}
\usepackage{bm}
\usepackage{graphicx}
\usepackage[dvipsnames]{xcolor}
\usepackage{booktabs}
\usepackage{subfigure}
\usepackage{multirow}

\newcommand{\RR}[2]{\textcolor[rgb]{0,0,0}
{#2}}

\ifCLASSINFOpdf
\else
\fi

\begin{document}

\title{Vehicle Instance Segmentation from Aerial Image and Video Using a Multi-Task Learning Residual Fully Convolutional Network}
\author{Lichao~Mou,~\IEEEmembership{Student Member,~IEEE,} and
       ~Xiao~Xiang~Zhu,~\IEEEmembership{Senior Member,~IEEE}
\thanks{This work is jointly supported by the China Scholarship Council, the European Research Council (ERC) under the European Union's Horizon 2020 research and innovation programme (grant agreement No. [ERC-2016-StG-714087], Acronym: \textit{So2Sat}), Helmholtz Association under the framework of the Young Investigators Group ``SiPEO'' (VH-NG-1018, www.sipeo.bgu.tum.de), and the Bavarian Academy of Sciences and Humanities in the framework of Junges Kolleg. }
\thanks{(\textit{Corresponding Author: Xiao Xiang Zhu})}

\thanks{The authors are with the Remote Sensing Technology Institute (IMF), German Aerospace Center (DLR), Germany and with Signal Processing in Earth Observation (SiPEO), Technical University of Munich (TUM), Germany (e-mails: lichao.mou@dlr.de; xiao.zhu@dlr.de).}
       }

\markboth{,IEEE Transactions on Geoscience and Remote Sensing, in press}%
{Shell }

\maketitle

\begin{abstract}

\textit{\textcolor{blue}{This is a preprint, to read the final version please go to IEEE Transactions on Geoscience and Remote Sensing on IEEE XPlore.}}

Object detection and semantic segmentation are two main themes in object retrieval from high-resolution remote sensing images, which have recently achieved remarkable performance by surfing the wave of deep learning and, more notably, convolutional neural networks (CNNs). In this paper, we are interested in a novel, more challenging problem of vehicle instance segmentation, which entails identifying, at a \RR{}{pixel-level}, where the vehicles appear as well as associating each pixel with a physical instance of a vehicle. In contrast, vehicle detection and semantic segmentation each only concern one of the two. We propose to tackle this problem with a semantic boundary-aware multi-task learning network. More specifically, we utilize the philosophy of residual learning (ResNet) to construct a fully convolutional network that is capable of harnessing multi-level contextual feature representations learned from different residual blocks. We theoretically analyze and discuss why residual networks can produce better probability maps for pixel-wise segmentation tasks. Then, based on this network architecture, we propose a unified multi-task learning network that can simultaneously learn two complementary tasks -- namely, segmenting vehicle regions and detecting semantic boundaries. The latter subproblem is helpful for differentiating ``touching'' vehicles, which are usually not correctly separated into instances. Currently, datasets with pixel-wise annotation for vehicle extraction are ISPRS dataset and IEEE GRSS DFC2015 dataset over Zeebrugge, which specializes in semantic segmentation. Therefore, we built a new, more challenging dataset for vehicle instance segmentation, called the \emph{Busy Parking Lot UAV Video dataset}, and \RR{}{we make our dataset available at \url{http://www.sipeo.bgu.tum.de/downloads}} so that it can be used to benchmark future vehicle instance segmentation algorithms.
\end{abstract}


\IEEEpeerreviewmaketitle

\section{Introduction}
\label{sec:intro}
The last decade has witnessed dramatic progress in modern remote sensing technologies -- along with the launch of small and cheap commercial high-resolution satellites and the now widespread availability of unmanned aerial vehicles (UAVs) -- which facilitates a diversity of applications, such as urban management~\cite{Volpi17,dfc16joint,Audebert17,rifcn}, monitoring of land changes~\cite{Vakalopoulou16,Wen16,Wu17,Lyu16}, and traffic monitoring~\cite{MouDFC16,Kopsiaftis15}. Among these applications, object extraction from very high-resolution remote sensing images/videos has gained increasing attention in the remote sensing community in recent years, particularly vehicle extraction, due to successful civil applications. Vehicle extraction, however, is still a challenging task, mainly because it is easily affected by several factors, e.g., vehicle appearance variation, the effects of shadow, illumination, a complicated and cluttered background, etc. Existing vehicle extraction approaches can be roughly divided into two categories: vehicle detection and vehicle semantic segmentation.

\begin{figure}[t]
\centering
\includegraphics[width=0.95\columnwidth]{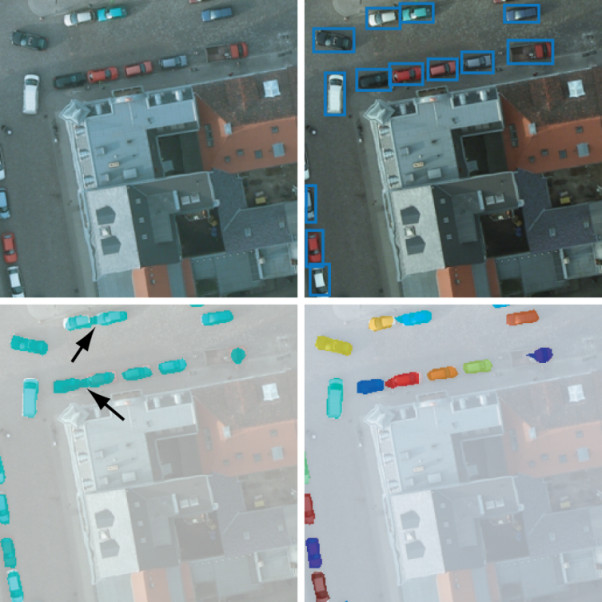}
\renewcommand{\figurename}{Fig}
\caption{\label{fig:intro} An illustration of different vehicle extraction methods. From left to right and top to bottom: input image, vehicle detection, semantic segmentation, and vehicle instance segmentation. The challenge of vehicle instance segmentation is that some vehicles are segmented incorrectly. While most pixels belonging to the category are identified correctly, they are not correctly separated into instances (see arrows in the lower left image).}
\end{figure}

\subsection{Vehicle Detection}
The goal of vehicle detection is to detect all instances of vehicles and localize them in the image, typically in the form of bounding boxes with confidence scores. Traditionally, this topic was addressed by works that use low-level, hand-crafted visual features (e.g., color histogram, texture feature, scale-invariant feature transform (SIFT), and histogram of oriented gradients (HOG)) and classifiers. For example, in~\cite{Shao12}, the authors incorporate multiple visual features, local binary pattern (LBP), HOG, and opponent histogram for vehicle detection from high-resolution aerial images. Moranduzzo and Melgani~\cite{Moranduzzo141} first use SIFT to detect the interest points of vehicles and then train a support vector machine (SVM) to classify these interest points into vehicle and non-vehicle categories based on the SIFT descriptors. They later present an approach~\cite{Moranduzzo142} that performs filtering operations in horizontal and vertical directions to extract HOG features and yield vehicle detection after the computation of a similarity measure, using a catalog of vehicles as a reference. \cite{KangLiu15}, where authors make use of an integral channel concept, with Haar-like features and an AdaBoost classifier in a soft-cascade structure, to achieve fast and robust vehicle detection.
\par
The aforementioned approaches mainly rely on hand-crafted features for constructing a classification system. Recently, as an important branch of the deep learning family, the convolutional neural network (CNN) has become the method of choice in many computer vision and remote sensing problems~\cite{Zhu17DLinRS,im2height,resconvdeconv,recnn,Mournn} (e.g., object detection) due to its ability to automatically extract mid- and high-level abstract features from raw images for pattern recognition purposes. Chen et al.~\cite{PanLetter14} propose a vehicle detection model, called the hybrid deep neural network, which consists of a sliding window technique and CNN. The main insight behind their model is to divide the feature maps of the last convolutional layer into different scales, allowing for the extraction of multi-scale features for vehicle detection. In~\cite{Ammour17}, authors segment an input image into homogeneous superpixels that can be considered as vehicle candidate regions, making use of a pre-trained deep CNN to extract features, and train a linear SVM to classify these candidate regions into vehicle and non-vehicle classes.

\subsection{Vehicle Semantic Segmentation}
Vehicle semantic segmentation aims to label each pixel in an image as belonging to the vehicle class or other categories (e.g., building, tree, low vegetation, etc.). In comparison with vehicle detection, it can give more accurate pixel-wise extraction results. More recently, progress in deep CNNs, particularly fully convolutional networks (FCNs), makes it possible to achieve end-to-end vehicle semantic segmentation. For instance, Audebert et al.~\cite{AudebertRS17} propose a deep learning-based ``segment-before-detect'' method for semantic segmentation and subsequent classification of several types of vehicles in high-resolution remote sensing images. The use of SegNet~\cite{SegNet} in this method is capable of producing pixel-wise annotations for vehicle semantic mapping. In addition, several recent works in semantic segmentation of high-resolution aerial imaging also involve vehicle segmentation. In~\cite{Kampffmeyer16}, the authors focus on class imbalance, which often represents a problem for semantic segmentation in remote sensing images since small objects (e.g., vehicles) are less prioritized in an effort to achieve good overall accuracy. To address this problem, they train FCNs, using the cross-entropy loss function weighted with median frequency balancing, which is proposed by Eigen and Fergus~\cite{Eigen15}.

\subsection{Is Semantic Segmentation Good Enough for Vehicle Extraction?}
The existence of ``touching'' vehicles in a remote sensing image makes it quite hard for most vehicle semantic segmentation methods to separate objects individually, while in most cases, we need to know not only which pixels belong to vehicles (vehicle semantic segmentation problem) but also the exact number of vehicles (vehicle detection task). This drives us to examine instance-oriented vehicle segmentation.
\par
Vehicle instance segmentation seeks to identify the semantic class of each pixel (i.e., vehicle or non-vehicle) as well as associate each pixel with a physical instance of a vehicle. This is contrasted with vehicle semantic segmentation, which is only concerned with the above-mentioned first task. In this work, we are interested in vehicle instance segmentation in a complex, cluttered, and challenging background from aerial images and videos. Moreover, since deep networks have recently been very successful in a variety of remote sensing applications, from hyper/multi-spectral image analysis to interpretation of high-resolution aerial images to multimodal data fusion~\cite{Zhu17DLinRS}, in this paper, we would like to use an end-to-end network to achieve vehicle instance segmentation. Our work contributes to the literature in three major respects:
\begin{itemize}
  \item So far, most studies in the remote sensing community have focused on object detection and semantic segmentation in high-resolution remote sensing imagery. Instance segmentation has rarely been addressed. In a pioneer work moving from semantic segmentation to instance segmentation, Audebert et al.~\cite{AudebertRS17} developed a three-stage ��segment-before-detect�� framework. In this paper, we try to address the vehicle instance segmentation problem by a end-to-end learning framework.
  \item In order to facilitate progress in the field of vehicle instance segmentation in high-resolution aerial images/videos, we provide a new, challenging dataset that presents a high range of variation -- with a diversity of vehicle appearances, effects of shadow, a cluttered background, and extremely close vehicle distances -- for producing quantitative measurements and comparing among approaches.
  \item We present a semantic boundary-aware unified multi-task learning fully convolutional network, which is end-to-end trainable, for vehicle instance segmentation. Inspired by several recent works~\cite{Zhaocvpr17,Wu16,Laina16}, we exploit ResNet~\cite{ResNet} to construct the feature extractor of the whole network. In this paper, we theoretically analyze and discuss why residual networks can produce better probability maps for pixel-wise prediction tasks. The proposed multi-task learning network creates  two separate, yet identical branches to jointly optimize two complementary tasks -- namely, vehicle semantic segmentation and semantic boundary detection. The latter subproblem is beneficial for differentiating vehicles with an extremely close distance and further improving instance segmentation performance.
\end{itemize}
\par
The remainder of this paper is organized as follows. After the introductory Section~\ref{sec:intro}, detailing vehicle extraction from high-resolution remote sensing imagery, we enter Section~\ref{sec:method}, dedicated to the details of the proposed semantic boundary-aware multi-task learning network for vehicle instance segmentation. Section~\ref{sec:exp} then provides dataset information, the network setup, and experimental results and discussion. Finally, Section~\ref{sec:conc} concludes the paper.

\section{Methodology}
\label{sec:method}
We formulate the vehicle instance segmentation task by two subproblems, namely vehicle detection and semantic segmentation. The training set is denoted by $\{(\bm{x}_i,\bm{y}_i,\bm{z}_i)\}$, where $i=1,2,\cdots,N$ and $N$ is the number of training samples. Since we consider each image independently, the subscript $i$ is dropped hereafter for notational simplicity. $\bm{x}=\{x_j,j=1,2,\cdots,|\bm{x}|\}$ represents a raw input image, $\bm{y}=\{y_j,j=1,2,\cdots,|\bm{x}|,y_j\in\{0,1\}\}$ denotes its corresponding manually annotated pixel-wise segmentation mask, and $\bm{z}=\{\bm{r}_k,k=0,1,\cdots,K\}$ is the instance label, where $\bm{r}_k$ indicates a set of pixels inside the $k$-th region\footnote{Regions in the image satisfy $\bm{r}_k\cap\bm{r}_t=\varnothing,\forall k\neq t$ and $\cup \bm{r}_k=\Omega$, in where $\Omega$ is the whole image region.}. $K$ is the total number of vehicle instances in the image, and $\bm{r}_0$ is the background area. When $k$ takes other values, it denotes the corresponding vehicle instance. Note that instance labels only count vehicle instances, thus they are commutative. Our aim is to segment vehicles while ensuring that all instances are differentiated. In this work, we approximate vehicle detection by semantic boundary detection\footnote{Semantic boundary detection is to detect the boundaries of each object instance in the images. Compared to edge detection, it focuses more on the association of boundaries and their object instances.}. We generate semantic boundary labels $\bm{b}$ through $\bm{z}$ to train a boundary detector, in which $\bm{b}=\{b_j,j=1,2,\cdots,|\bm{x}|,b_j\in\{0,1\}\}$ and $b_j$ equals 1 when it belongs to boundaries.
\par
In this section, we describe in detail our proposed semantic boundary-aware multi-task learning network for accurate vehicle instance segmentation. We start by introducing the FCN architecture for end-to-end semantic segmentation in Section~\ref{sec:a}. Furthermore, we propose to exploit multi-level contextual feature representations, generated by different stages of a residual network, to construct a residual FCN for producing better likelihood maps of vehicle regions or semantic boundaries (see Section~\ref{sec:b}). Then, in Section~\ref{sec:c}, we elaborate the semantic boundary-aware unified multi-task learning network drawn from the residual FCN for effective instance segmentation by jointly optimizing the complementary tasks.

\begin{figure*}[t]
\centering
\includegraphics[width=\linewidth]{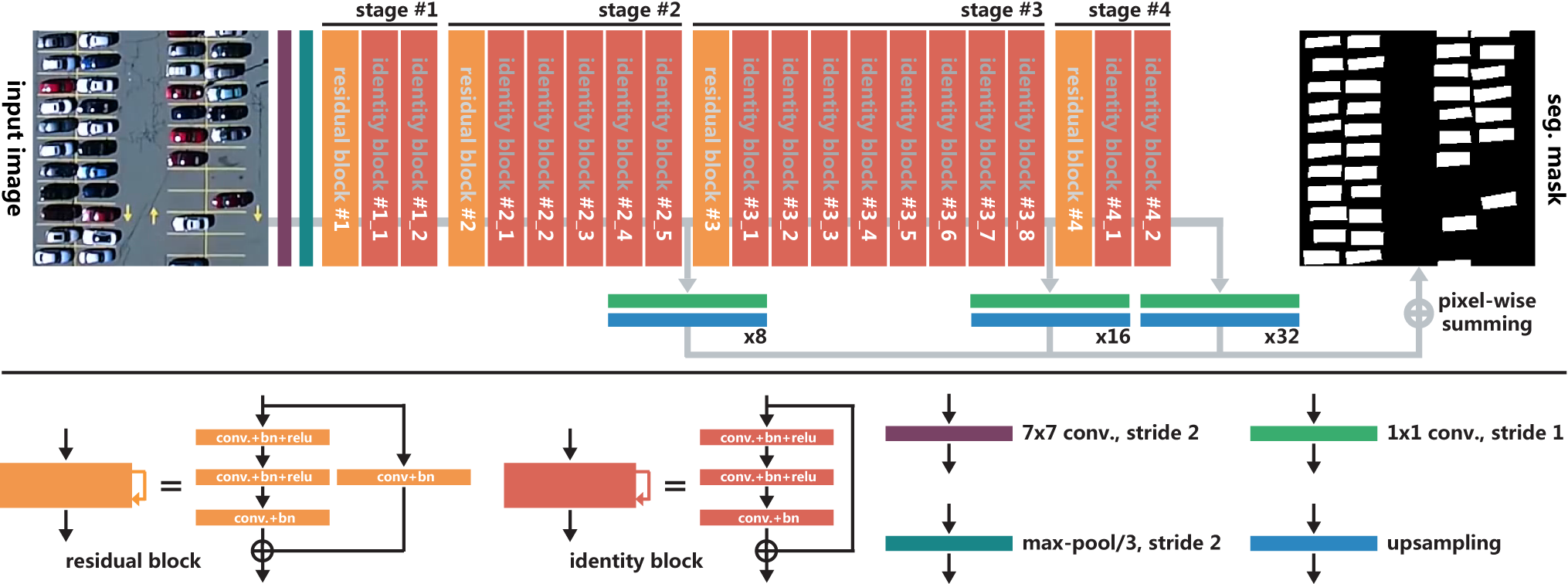}
\renewcommand{\figurename}{Fig}
\caption{\label{fig:network1} The network architecture of the ResFCN we use, as illustrated in Section~\ref{sec:b}. We incorporate multi-level contextual features from the last $32\times32$, $16\times16$, and $8\times8$ layers of a classification ResNet since making use of information from fairly early fine-grained layers is beneficial to segmenting small objects such as vehicles. To get the desired full resolution output, we use $1\times1$ convolutional layers followed by upsampling operations to upsample back to the spatial resolution of the input image. Then, predictions from different residual blocks are fused together with a summing operation.}
\end{figure*}

\subsection{Fully Convolutional Network for Semantic Segmentation}
\label{sec:a}
Long et al.~\cite{FCN} first proposed FCN architecture for semantic segmentation tasks, which is both efficient and effective. Later, some extensions of the FCN model have been proposed to improve semantic segmentation performance. To name a few, in~\cite{deepLab}, the authors removed some of the max-pooling operations and, accordingly, introduced atrous/dilated convolutions in their network, which can expand the field of view without increasing the number of parameters. As post-processing, a dense conditional random field (CRF) was trained separately to refine the estimated category score maps for further improvement. Zhang et al.~\cite{CRF-RNN} introduced a new form of network that combines FCN- and CRF-based probabilistic graphical modeling to simulate mean-field approximate inference for the CRF, with Gaussian pairwise potentials as the recurrent neural network (RNN).

\subsection{Residual Fully Convolutional Network (ResFCN)}
\label{sec:b}
Here, we first explain how to construct a ResFCN according to existing works in literature; mainly, the ResNet~\cite{ResNet} and FCN~\cite{FCN}. Then, we theoretically analyze why ResFCN is able to offer better performance than other FCNs based on the traditional feedforward network architectures (e.g., VGG Nets~\cite{VGGNet}).
\par
\textbf{Network design.} Several recent studies in computer vision~\cite{Zhaocvpr17,Wu16,Laina16} have shown that ResNet~\cite{ResNet} is capable of offering better features for pixel-wise prediction tasks such as semantic segmentation~\cite{Zhaocvpr17,Wu16} and depth estimation~\cite{Laina16}. We, therefore, make use of ResNet to construct the segmentation network in our work. We initialize a ResFCN from the original version of ResNet~\cite{ResNet}, instead of the newly presented pre-activation version~\cite{ResNet2}. Unlike~\cite{FCN}, we directly remove the fully connected layers from the original ResNet but do not convolutionalize these layers so as to make one prediction per spatial location. Moreover, we keep the $7\times7$ convolutional layer and $3\times3$ max-pooling layer, which can enlarge the field of view for feature representations. One of recent trend in network architecture design is stacking convolutional layers with small convolution kernels (e.g., $3\times3$ and $1\times1$) in the entire network because the stacked small kernels are more efficient than a large filter, given the same computational complexity. However, a recent study~\cite{LargeKernel} found that the large filter also plays an important role when classification and localization tasks are performed simultaneously. This can be easily understood through the analogy of individuals commonly confirming the category of a pixel by referring to its surrounding context region.
\par
By now, the output feature maps are only $\left. 1 \middle/ 32 \right.$ the resolution of their original input image, which is apparently too low to precisely differentiate individual pixels. To deal with this problem, Long et al.~\cite{FCN} made use of backwards strided convolutions that upsample the feature maps and output score masks. The motivation behind this is that the convolutional layers and max-pooling layers focus on extracting high-level abstract features whereas the backwards strided convolutions estimate the score masks in a pixel-wise way. Ghiasi et al.~\cite{Ghiasi16} proposed a multi-resolution reconstruction architecture based on a Laplacian pyramid that uses skip connections from higher resolution feature maps and multiplicative gating to successively refine segment boundaries reconstructed from lower-resolution maps. Inspired by the existing works, in this paper, we exploit multi-level contextual feature representations that include information from different residual blocks (i.e., different levels of contextual information). Fig.~\ref{fig:network1} shows the illustration of the ResFCN architecture we use with multi-level contextual features. More specifically, we incorporate feature representations from the last $32\times32$, $16\times16$, and $8\times8$ layers of the original ResNet since making use of information from fairly early fine-grained layers is beneficial to segmenting small objects such as vehicles. To get the desired full resolution output, we used a $1\times1$ convolutional layer, which adaptively squashes the number of channels down to the number of labels (1 for binary classification), takes advantage of the upsampling operation to upsample back to the spatial resolution of the input image, and makes predictions based on contextual cues from the given fields of view. Then, these predictions are fused together with a summing operation, and the final segmentation results are generated after sigmoid classification.

\begin{figure*}[t]
\centering
\includegraphics[width=0.9\linewidth]{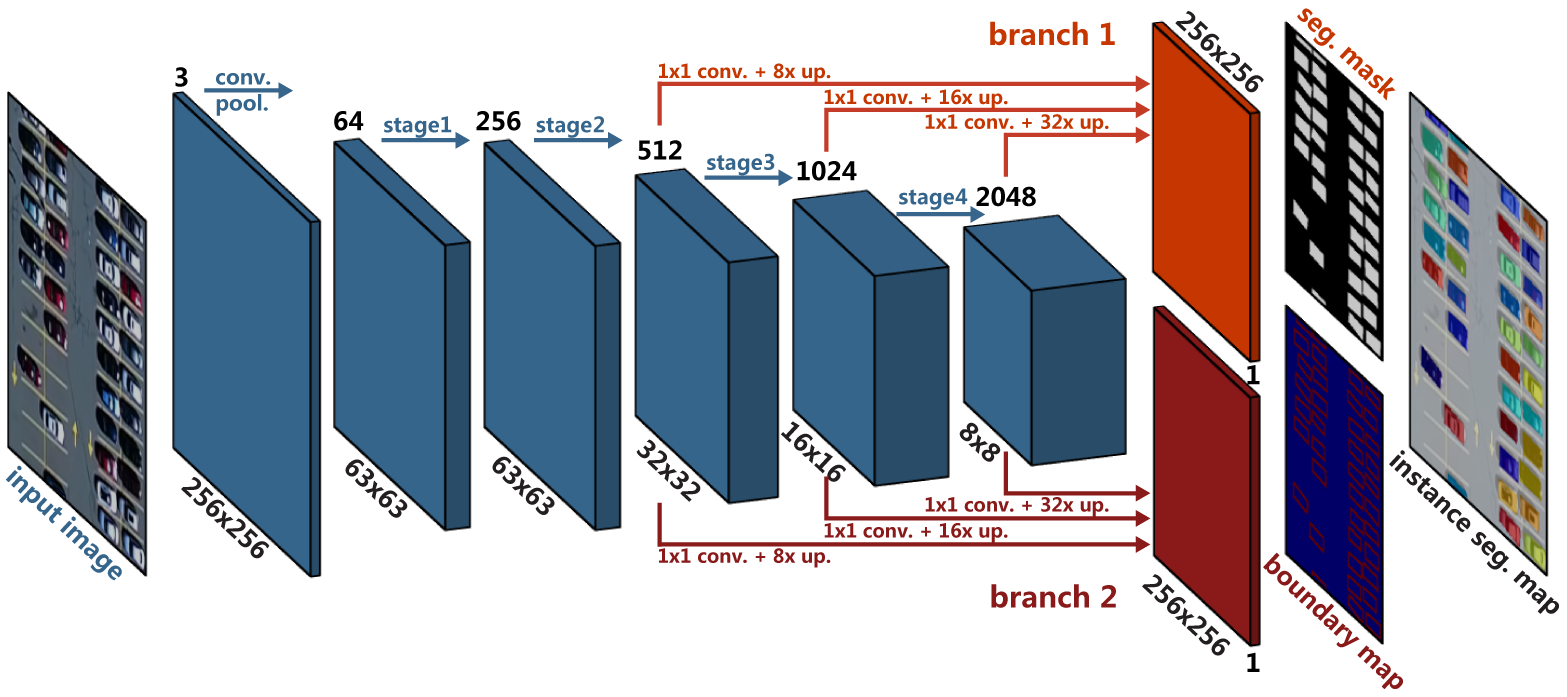}
\renewcommand{\figurename}{Fig}
\caption{\label{fig:network2} Overall architecture of the proposed semantic boundary-aware ResFCN. We propose to use such a unified multi-task learning network for vehicle instance segmentation, which creates two separate, yet identical branches to jointly optimize two complementary tasks, namely, vehicle semantic segmentation and semantic boundary detection. The latter subproblem is beneficial for differentiating ``touching'' vehicles and further improving the instance segmentation performance.}
\end{figure*}

\par
\textbf{Why residual learning?} Until recently, the majority of feedforward networks, like AlexNet~\cite{AlexNet} and VGG Nets~\cite{VGGNet}, were made up of a linear sequence of layers. $\bm{x}_{n-1}$ and $\bm{x}_n$ are denoted as the input and output of the $n$-th layer/block, respectively, and each layer in such a network learns the mapping function $\mathcal{F}$:
\begin{equation}\label{eq:res1}
\bm{x}_n=\mathcal{F}(\bm{x}_{n-1};\bm{\Theta}_n)\,,
\end{equation}
where $\bm{\Theta}_n$ is the parameters of the $n$-th layer. This kind of network is also often referred to as a traditional feedforward network.
\par
According to a study by He et al.~\cite{ResNet}, simply deepening traditional feedforward networks usually leads to an increase in training and test errors (i.e., so-called degradation problem). A residual learning-based network is composed of a sequence of residual blocks and exhibits significantly improved training characteristics, providing the opportunity to make network depths that were previously unattainable.
The output $\bm{x}_n$ of the $n$-th residual block in a ResNet can be computed as
\begin{equation}\label{eq:res2}
\bm{x}_n=\mathcal{H}(\bm{x}_{n-1};\bm{\Theta}_n)+\bm{x}_{n-1}\,,
\end{equation}
where $\mathcal{H}(\bm{x}_{n-1};\bm{\Theta}_n)$ is the residual, which is parametrized by $\bm{\Theta}_n$. The core insight of ResNet is that the addition of a shortcut connection from the input $\bm{x}_{n-1}$ to the output $\bm{x}_n$ bypasses two or more convolutional layers by performing identity mapping and is then added together with the output of stacked convolutions. By doing so, $\mathcal{H}$ only computes a residual instead of computing the output $\bm{x}_n$ directly.
\par
In the experiments, we found that ResFCN can offer better performance than other FCNs based on traditional feedforward network architecture, such as VGG-FCN. What is the reason behind this? To answer this question, we need to go deeper. According to the characteristics of ResFCN, we can easily get the following recurrence formula
\begin{equation}\label{eq:res3}
\bm{x}_m=\sum_{i=n-1}^{m-1}\mathcal{H}(\bm{x}_{i};\bm{\Theta}_{i+1})+\bm{x}_{n-1}\,,
\end{equation}
for any deeper residual block $m$ and any shallower residual block $n$. Eq. (\ref{eq:res3}) shows that ResFCN creates a direct path for propagating information of shallow layers (i.e., $\bm{x}_{n-1}$) through the entire network. Several recent studies~\cite{Zeiler14,Mahendran15} that attempt to reveal what were learned by CNNs show that deeper layers exploit filters to grasp global high-level information while shallower layers capture low-level details, such as object boundaries and edges, which are of great importance in small object detection/segmentation. In addition, when we dive into the backward propagation process, according to the chain rule of backpropagation, we can obtain
\begin{equation}\label{eq:res4}
\begin{split}
\frac{\partial \mathcal{E}}{\partial \bm{x}_{n-1}}&=\frac{\partial \mathcal{E}}{\partial \bm{x}_m}\frac{\partial \bm{x}_m}{\partial \bm{x}_{n-1}}\\
&=\frac{\partial \mathcal{E}}{\partial \bm{x}_m}(1+\frac{\partial}{\partial \bm{x}_{n-1}}\sum_{i=n-1}^{m-1}\mathcal{H}(\bm{x}_{i};\bm{\Theta}_{i+1}))\,,
\end{split}
\end{equation}
where $\mathcal{E}$ is the loss function of the network. As exhibited in Eq. (\ref{eq:res4}), the gradient $\frac{\partial \mathcal{E}}{\partial \bm{x}_{n-1}}$ can be decomposed into two additive terms: the term $\frac{\partial \mathcal{E}}{\partial \bm{x}_m}(\frac{\partial}{\partial \bm{x}_{n-1}}\sum_{i=n-1}^{m-1}\mathcal{H})$ that passes information through the weight layers, and the term $\frac{\partial \mathcal{E}}{\partial \bm{x}_m}$ that directly propagates without concerning any weight layers. The latter term ensures that the information can also be directly propagated back to any shallower residual block $n$.
\par
In brief, the properties of the forward and backward propagation procedures of the ResFCN make it possible to shuttle the low-level visual information directly across the network, which is quite helpful for our vehicle (small object) instance segmentation tasks.

\subsection{Semantic Boundary-Aware ResFCN}
\label{sec:c}
By exploiting the multi-level contextual features, the ResFCN is capable of producing good likelihood maps of vehicles. It is, however, still difficult to differentiate vehicles with a very close distance by only leveraging the probability of vehicles, due to the ambiguity in ``touching'' regions. This is rooted in the loss of spatial details caused by max-pooling layers (downsampling) along with feature abstraction. The semantic boundaries of vehicles provide good complementary cues that can be used for separating instances.
\par
Some approaches in computer vision and remote sensing have been explored for modeling segmentation and boundary prediction jointly in a combinatorial framework. For example, Kirillov et al.~\cite{Kirillov17} propose InstanceCut, which represents instance segmentation by two modalities, namely a semantic segmentation and all instance-boundaries. The former is computed from a CNN for semantic segmentation, and the latter is derived from a instance-aware edge detector. But this approach does not address end-to-end learning. In the remote sensing community, Marmanis et al.~\cite{Marmanis18} propose a two-step model that learns a CNN to separately output edge likelihoods at multiple scales from color-infrared (CIR) and height data. Then, the boundaries detected with each source are added as an extra channel to each source, and a network is trained for semantic segmentation purposes. The intuition behind this work is that using predicted boundaries helps to achieve sharper segmentation maps. In contrast, we train one end-to-end network that takes as input color images and predicts segmentation maps and object boundaries, in order to augment the performance of segmentation at instance level.
\par
To this end, we train a deep semantic boundary-aware ResFCN for effective vehicle instance segmentation (i.e., segmenting the vehicles and splitting clustered instances into individual ones). Fig.~\ref{fig:network2} shows an overview of the proposed network. Specifically, we formulate it as a unified multi-task learning network architecture by exploring the complementary information (i.e., vehicle region and semantic boundaries), instead of treating the vehicle segmentation problem as an independent and single task, which can simultaneously learn the detections of vehicle regions and corresponding semantic boundaries. As shown in Fig.~\ref{fig:network2}, the feature representations extracted from multiple residual blocks are upsampled with two separate, yet identical branches to predict the semantic segmentation masks of vehicles and semantic boundaries, respectively. In each branch, the mask is estimated by the ResFCN with multi-level contextual features as illustrated in Section~\ref{sec:b}. Since we have only two categories (foreground/vehicles vs. background and semantic boundaries vs. non-boundaries), sigmoid and binary cross-entropy loss are used to train these two branches. Formally, the network training can be formulated as a pixel-level binary classification problem regarding ground truth segmentation masks, including vehicle instances and semantic boundaries, as shown in the following:
\begin{equation}\label{eq:mtl1}
\mathcal{L}(x;\bm{W})=\mathcal{L}_s(x;\bm{W}_n,\bm{W}_s)+\lambda \mathcal{L}_b(x;\bm{W}_n,\bm{W}_b)\,,
\end{equation}
where
\begin{equation}\label{eq:mtl1}
\begin{split}
&\mathcal{L}_s=-\sum_{x\in \bm{x}}[y\log \sigma_s(x)+(1-y)\log (1-\sigma_s(x))]\,,\\
&\mathcal{L}_b=-\sum_{x\in \bm{x}}[b\log \sigma_b(x)+(1-b)\log (1-\sigma_b(x))]\,.
\end{split}
\end{equation}
\par
$\mathcal{L}_s(x;\bm{W}_n,\bm{W}_s)$ and $\mathcal{L}_b(x;\bm{W}_n,\bm{W}_s)$ denote losses for estimating vehicle regions and semantic boundaries, respectively. We train the network using this joint loss, and the final instance segmentation map is produced by the first branch of the network in test phase. Vehicle instances are obtained by computing connected regions in the predicted segmentation map. Inside a region, pixels belong to the same vehicle; while different regions mean different instances. Our motivation is that jointly estimating segmentation and boundary map in a multi-task network with such a joint loss can offer a better segmentation result at instance level for aerial images. Note that we do not make use of any post-processing operations, such as fusing the segmentation and boundary map, as we want to directly evaluate the performance of this network architecture.
\par
Note that the multi-task learning network is optimized in an end-to-end fashion. This joint multi-task training procedure has several merits. First, in the application of vehicle instance segmentation, the multi-task learning network architecture is able to provide complementary semantic boundary information, which is helpful in differentiating the clustered vehicles, improving the instance-level segmentation performance. Second, the discriminative capability of the network's intermediate feature representations can be improved by this architecture because of multiple regularizations on correlated tasks. Therefore, it can increase the robustness of instance segmentation performance.

\section{Experimental Results and Discussion}
\label{sec:exp}
\subsection{Datasets}
\subsubsection{\textbf{ISPRS Potsdam}}
\begin{figure}[t]
\centering
\includegraphics[width=\columnwidth]{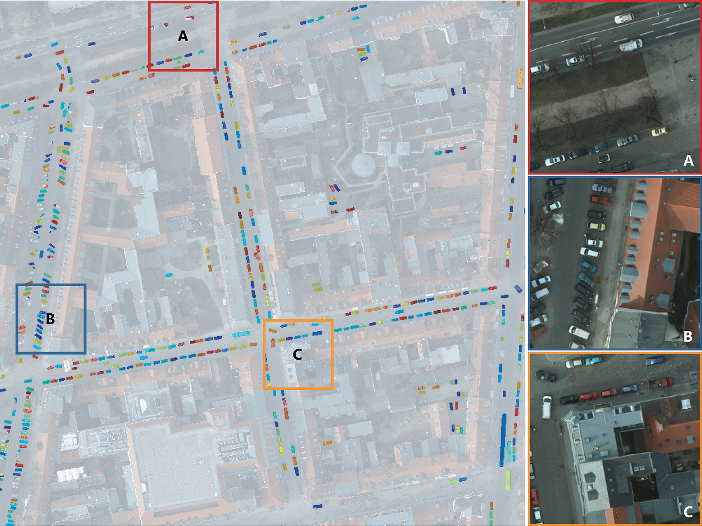}
\renewcommand{\figurename}{Fig}
\caption{\label{fig:potsdam} Image \#5\_12 from the ISPRS Potsdam dataset for vehicle instance segmentation as well as three zoomed-in areas.}
\end{figure}

The ISPRS Potsdam Semantic Labeling dataset~\cite{potsdam} is an open benchmark dataset provided online\footnote{\url{http://www2.isprs.org/commissions/comm3/wg4/2d-sem-label-potsdam.html}}. The dataset is consists of 38 ortho-rectified aerial IRRGB images ($6000\times6000$ px), with a 5 cm spatial resolution and corresponding DSMs generated by dense image matching, taken over the city of Potsdam, Germany. A comprehensive manually annotated pixel-wise segmentation mask is provided as ground truth for 24 tiles, which are available for training and validation. The other 14 remain unreleased and are kept with the challenge organizers for testing purposes. We randomly selected 5 tiles (image number: 2\_12, 5\_12, 7\_7, 7\_8, 7\_9) from 24 training images and used them as test set in our experiments (cf. Fig.~\ref{fig:potsdam}). The resolution is downsampled to 15 cm/pixel to match the subsequent video dataset. The input to the networks contains only red, green, and blue channels, and all results reported on this dataset refer to the aforementioned test set. Table~\ref{tab:potsdam} provides details about this dataset for our experiments.

\subsubsection{\textbf{Busy Parking Lot}}
\begin{figure}[t]
\centering
\includegraphics[width=\columnwidth]{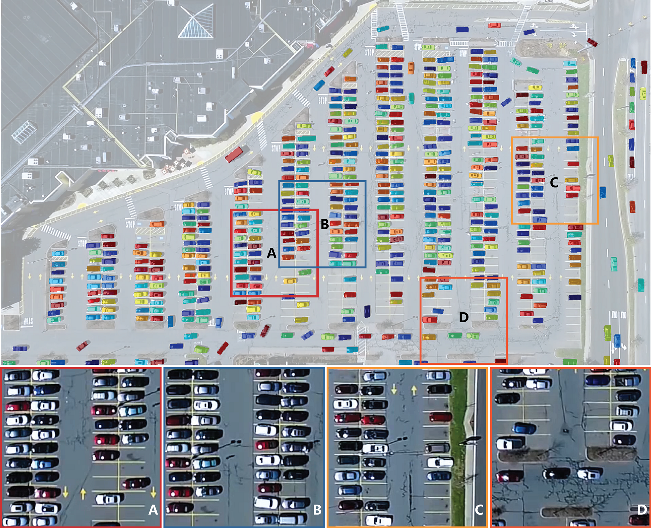}
\renewcommand{\figurename}{Fig}
\caption{\label{fig:parkinglot} Frame@1s from the proposed Busy Parking Lot UAV Video dataset for vehicle instance segmentation. Four zoomed-in areas are shown on the bottom.}
\end{figure}

The task of vehicle instance segmentation currently lacks a compelling and challenging benchmark dataset to produce quantitative measurements and to compare with other approaches. While the ISPRS Potsdam dataset has clearly boosted research in semantic segmentation of high-resolution aerial imagery, it is not as challenging as certain practical scenes, such as a busy parking lot, where vehicles are often parked so close that it is quite hard to separate them, particularly from an aerial view. To this end, in this work, we propose our new challenging Busy Parking Lot UAV Video dataset that we built for the vehicle instance segmentation task.
The UAV video was acquired by a camera onboard a UAV covering the parking lot of Woburn Mall, in Woburn, Massachusetts, USA. The video comprises $1920\times1080$ pixels with a spatial resolution of about 15 cm per pixel at 24 frames per second and with a length of 60 seconds. We have manually annotated pixel-wise instance segmentation masks for 5 frames (at 1, 15, 30, 45, and 59 seconds); i.e., the annotation is dense in space and sparse in time to allow for the evaluation of methods with this long sequence (cf. Fig.~\ref{fig:parkinglot}). The Busy Parking Lot dataset is challenging because it presents a high range of variations, with a diversity of vehicle colors, effects of shadow, several slightly blurred regions, and vehicles that are parked too close. We train networks on the ISPRS Potsdam dataset and then perform vehicle instance segmentation using the trained networks on this video dataset. Details regarding this dataset are shown in Table~\ref{tab:parkinglot}.

\begin{table*}[t]
\caption{\label{tab:potsdam} Vehicle Counts and Number of Vehicle Pixels in ISPRS Potsdam Dataset}
\centering
\begin{tabular}{ccccccc}
\toprule
 & \multirow{2}{*}{\textbf{Training Set}} & \multicolumn{5}{c}{\textbf{Test Set}} \\
\cline{3-7}
 & & \textbf{2\_12} & \textbf{5\_12} & \textbf{7\_7} & \textbf{7\_8} & \textbf{7\_9} \\
\hline
Vehicle Count & 4,433 & 123 & 427 & 301 & 309 & 305 \\
Number of Pixels & 1,184,789 & 36,236 & 122,332 & 76,892 & 77,669 & 74,404 \\
\bottomrule
\end{tabular}
\end{table*}

\begin{table*}[t]
\caption{\label{tab:parkinglot} Vehicle Counts and Number of Vehicle Pixels in Busy Parking Lot UAV Video Dataset}
\centering
\begin{tabular}{cccccc}
\toprule
 & \textbf{Frame@1s} & \textbf{Frame@15s} & \textbf{Frame@30s} & \textbf{Frame@45s} & \textbf{Frame@59s} \\
\hline
Vehicle Count & 511 & 492 & 502 & 484 & 479 \\
Number of Pixels & 257,462 & 235,560 & 240,607 & 235,448 & 226,697 \\
\bottomrule
\end{tabular}
\end{table*}

\subsection{Training Details}
\begin{figure}[ht]
\centering
\includegraphics[width=\columnwidth]{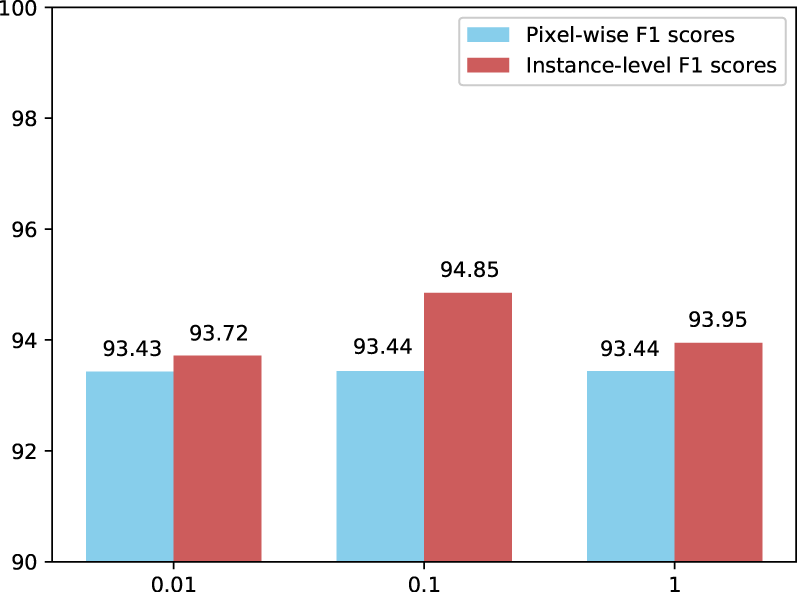}
\renewcommand{\figurename}{Fig}
\caption{\label{fig:lambda} \RR{}{A sensitivity analysis for the parameter $\lambda$ on ISPRS Potsdam dataset.}}
\end{figure}

The network training is based on the TensorFlow framework. We chose Nesterov Adam~\cite{nadam2,nadam1} as the optimizer to train the network, since, for this task, it shows much faster convergence than standard stochastic gradient descent (SGD) with momentum~\cite{sgd} or Adam~\cite{adam}. We fixed almost all of the parameters of Nesterov Aadam as recommended in~\cite{nadam2}: $\beta_1=0.9$, $\beta_2=0.999$, $\epsilon=1\mathrm{e}{-08}$, and a schedule decay of 0.004, making use of a fairly small learning rate of $2\mathrm{e}{-04}$. All weights in the newly added layers are initialized with a Glorot uniform initializer~\cite{Glorot_normal} that draws samples from a uniform distribution. \RR{}{In our experiments, we note that the pixel-wise F1 score of the network is less sensitive to the parameter $\lambda$, and the instance-level performance is relatively sensitive to $\lambda$. Based on the sensitivity analysis (cf. Fig.~\ref{fig:lambda}), we set it as 0.1.}

\par
The networks are trained on the training set of the ISPRS Potsdam dataset to predict instance segmentation maps. The training set has only 931 unique $256\times256$ patches. We make use of the data augmentation technique to increase the number of training samples. The RGB patches and corresponding pixel-wise ground truth are transformed by horizontally and vertically flipping three-quarters of the patches. By doing so, the number of training samples increases to 14,896. To monitor overfitting during training, we randomly select 10\% of the training samples as the validation set; i.e., splitting the training set into 13,406 training and 1,490 validation pairs. We train the network for 50 epochs and make use of early stopping to avoid overfitting. Moreover, we use fairly small mini-batches of 8 image pairs because, in a sense, every pixel is a training sample. We train our network on a single NVIDIA GeForce GTX TITAN with 12 GB of GPU memory, which takes about two hours.

\begin{figure*}[t]
\centering
\includegraphics[width=\linewidth]{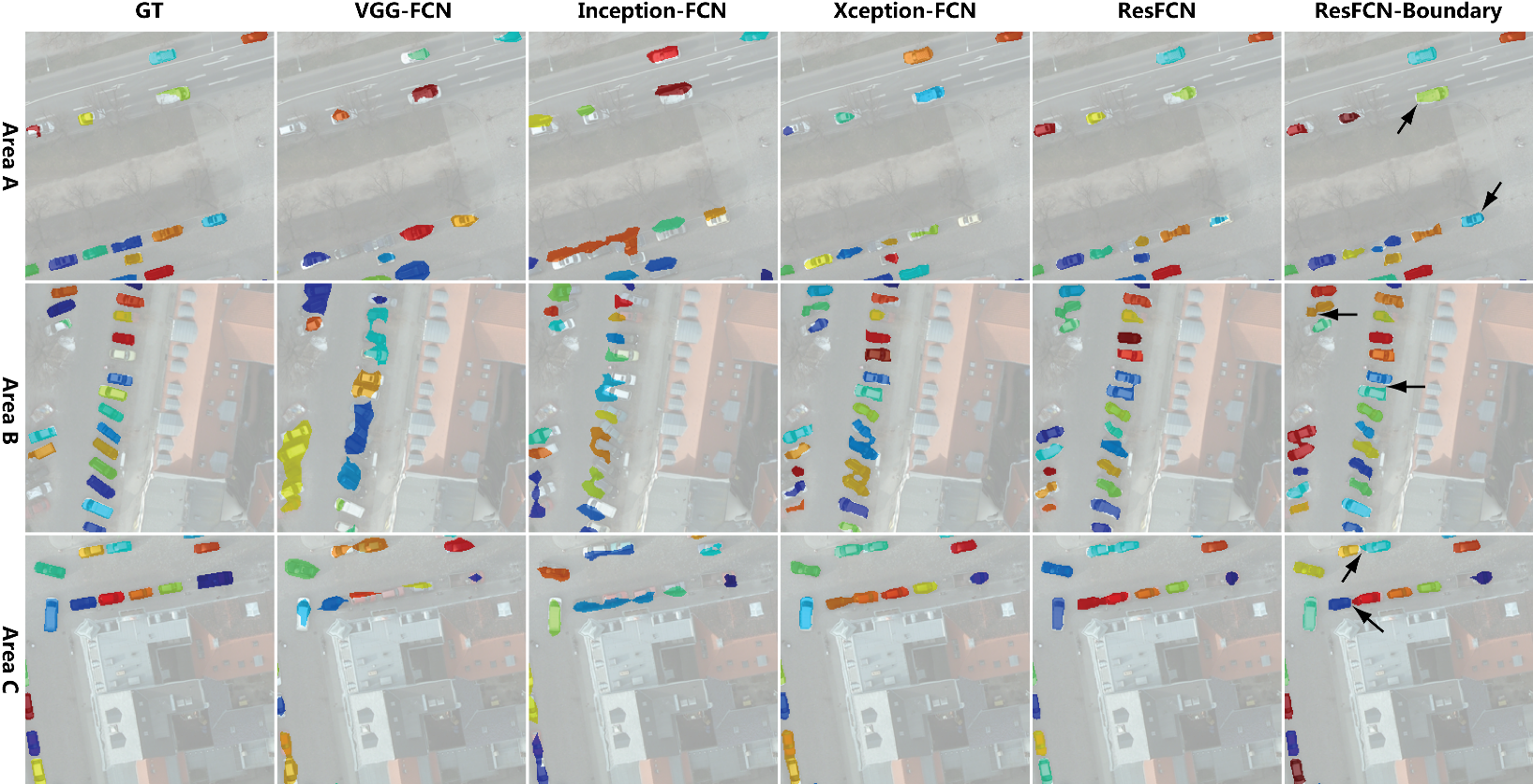}
\renewcommand{\figurename}{Fig}
\caption{\label{fig:potsdam_zoom} Instance segmentation results of ISPRS Potsdam dataset (from left to right): ground truth, VGG-FCN, Inception-FCN, Xception-FCN, ResFCN, and B-ResFCN (different colors denote individual vehicle objects). The three areas are derived from Fig.~\ref{fig:potsdam}.}
\end{figure*}

\begin{table}[t]
\caption{\label{tab:oa} \RR{}{Pixel-level OAs and F1-scores for the car class} on ISPRS Potsdam dataset}
\centering
\begin{tabular}{cccccc}
\toprule
\textbf{Model} & \textbf{OA} & \textbf{OA (eroded)} & \textbf{F1 score} & \textbf{F1 score (eroded)} \\
\hline
ResFCN & 99.79 & 99.89 & 93.43 & 95.66 \\
B-ResFCN & 99.79 & 99.89 & 93.44 & 95.87 \\
\bottomrule
\end{tabular}
\end{table}

\subsection{Qualitative Evaluation}
\begin{figure*}[!t]
\centering
\includegraphics[width=0.95\linewidth]{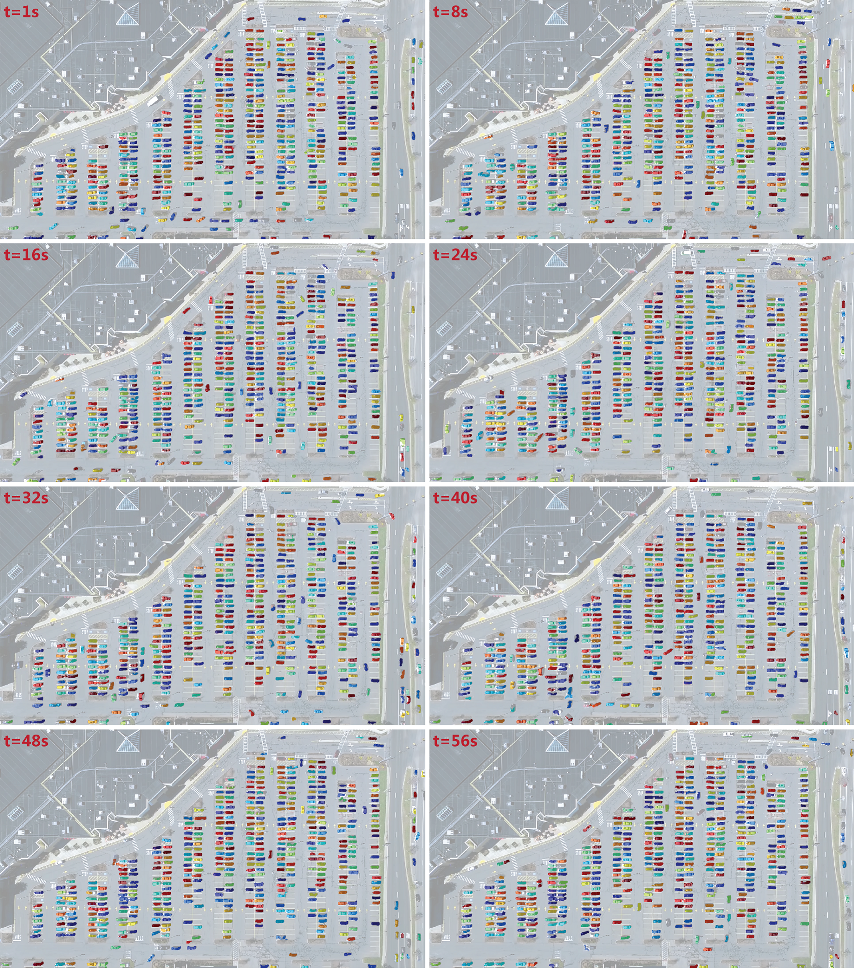}
\renewcommand{\figurename}{Fig}
\caption{\label{fig:parkinglot_frames} Example segmentations using the proposed B-ResFCN in several frames of the Busy Parking Lot dataset.}
\end{figure*}

\begin{figure*}[!t]
\centering
\includegraphics[width=0.85\linewidth]{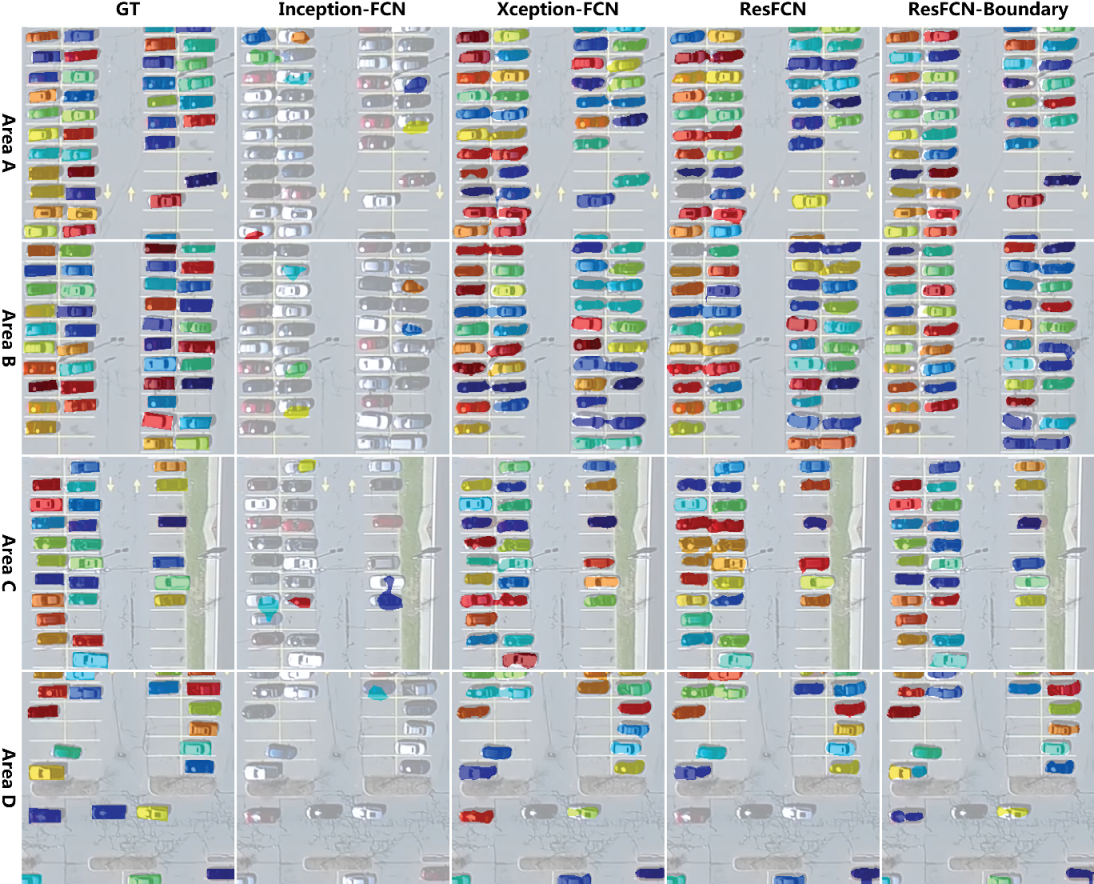}
\renewcommand{\figurename}{Fig}
\caption{\label{fig:parkinglot_zoom} Instance segmentation maps of Busy Parking Lot dataset (from left to right): ground truth, Inception-FCN, Xception-FCN, ResFCN, and B-ResFCN (different colors denote individual vehicles). The four areas are derived from Fig.~\ref{fig:parkinglot}.}
\end{figure*}

Some vehicle instance segmentation results are shown in Fig.~\ref{fig:potsdam_zoom} (test set of ISPRS Potsdam dataset) and Fig.~\ref{fig:parkinglot_zoom} (Busy Parking Lot dataset), respectively, in order to qualitatively illustrate the efficacy of our model. First, we compare various CNN variants used for FCN architecture to determine which one is the best-suited for our task. In Fig.~\ref{fig:potsdam_zoom}, we qualitatively investigate the accuracy of the predicted instance segmentation maps, using FCN architecture with leading CNN variants -- namely, VGG\cite{VGGNet}-FCN, Inception\cite{Szegedy16}-FCN, Xception\cite{Chollet17}-FCN, and ResFCN, on the ISPRS Potsdam dataset. We implement VGG-FCN, Inception-FCN, and Xception-FCN by fusing the output feature maps of the last three convolutional blocks as we do for ResFCN (cf. Section~\ref{sec:b}). From the segmentation results, we can see an improvement in quality from VGG-FCN to ResFCN. Moreover, on the Busy Parking Lot dataset, ResFCN also demonstrates a fairly strong ability to generalize to an ``unseen'' scene outside the training dataset (see Fig.~\ref{fig:parkinglot_zoom}). However, there are some vehicles that cannot be separated in both segmentation results produced using the aforementioned networks, due to the extremely close vehicle distance. The situation is further deteriorated when the imagery suffers from the effects of shadow, as the cases shown in the zoomed-in areas of Fig.~\ref{fig:parkinglot_zoom}. On the other hand, to identify the role of the semantic boundary component of the proposed unified multi-task learning network architecture, we also performed an ablation study to compare the performance of networks relying on the prediction of vehicles. In comparison with ResFCN, the semantic boundary-aware ResFCN (B-ResFCN) is able to separate those ``touching'' cars clearly, which qualitatively highlights the superiority of semantic boundary-aware network by exploring the complementary information under a unified multi-task learning network architecture. Fig.~\ref{fig:parkinglot_frames} shows a couple of example segmentations using the proposed B-ResFCN on several frames of the Busy Parking Lot dataset.

\begin{table*}[t]
\caption{\label{tab:det_potsdam} Detection Results of Different Networks on ISPRS Potsdam Semantic Labeling Dataset (Instance-level F1 Score, Precision, and Recall)}
\centering
\begin{tabular}{cccccccccccccccc}
\toprule
\multirow{2}{*}{\textbf{Model}} & \multicolumn{3}{c}{\textbf{2\_12}} & \multicolumn{3}{c}{\textbf{5\_12}} & \multicolumn{3}{c}{\textbf{7\_7}} & \multicolumn{3}{c}{\textbf{7\_8}} & \multicolumn{3}{c}{\textbf{7\_9}} \\
\cline{2-16}
 & \textbf{F1} & \textbf{P} & \textbf{R} & \textbf{F1} & \textbf{P} & \textbf{R} & \textbf{F1} & \textbf{P} & \textbf{R} & \textbf{F1} & \textbf{P} & \textbf{R} & \textbf{F1} & \textbf{P} & \textbf{R} \\
\hline
VGG-FCN & 66.04 & 70.00 & 62.50 & 57.00 & 61.45 & 53.14 & 59.21 & 61.95 & 56.70 & 57.21 & 66.84 & 50.00 & 61.31 & 65.91 & 57.31 \\
B-VGG-FCN & 70.27 & 68.42 & 72.22 & 69.85 & 67.42 & 72.47 & 71.03 & 68.47 & 73.79 & 67.96 & 66.86 & 69.09 & 66.47 & 60.96 & 73.08 \\
Inception-FCN & 51.91 & 55.45 & 48.80 & 31.65 & 37.42 & 27.42 & 40.00 & 43.41 & 37.08 & 27.79 & 31.70 & 24.74& 40.87 & 45.02 & 37.42 \\
B-Inception-FCN & 55.15 & 50.61 & 60.58 & 46.14 & 47.42 & 44.92 & 53.81 & 52.91 & 54.75 & 43.47 & 42.45 & 44.54 & 50.74 & 47.49 & 54.47 \\
Xception-FCN & 96.92 & 98.21 & 95.65 & 83.55 & 81.11 & 86.14 & 93.33 & 94.59 & 92.11 & 92.05 & 93.10 & 91.01 & 93.92 & 96.59 & 91.40 \\
B-Xception-FCN & 97.00 & \textbf{100} & 94.17 & 88.40 & \textbf{88.60} & 88.19 & 93.65 & 96.47 & 91.00 & 93.58 & 97.54 & 89.94 & 94.63 & 97.50 & 91.92 \\
ResFCN & 97.93 & \textbf{100} & 95.93 & 83.88 & 80.84 & 87.15 & 94.72 & 96.86 & 92.67 & \textbf{95.62} & \textbf{97.93} & \textbf{93.42} & 95.25 & 96.23 & \textbf{94.30} \\
B-ResFCN & \textbf{98.31} & \textbf{100} & \textbf{96.67} & \textbf{88.57} & 87.08 & \textbf{90.11} & \textbf{96.43} & \textbf{97.12} & \textbf{95.74} & 95.19 & 97.88 & 92.64 & \textbf{95.76} & \textbf{97.83} & 93.77 \\
\bottomrule
\end{tabular}
\end{table*}

\begin{table*}[t]
\caption{\label{tab:det_parkinglot} Detection Results of Different Methods on proposed Busy Parking Lot UAV Video Dataset (Instance-level F1 Score, Precision, and Recall)}
\centering
\begin{tabular}{cccccccccccccccc}
\toprule
\multirow{2}{*}{\textbf{Model}} & \multicolumn{3}{c}{\textbf{Frame@1s}} & \multicolumn{3}{c}{\textbf{Frame@15s}} & \multicolumn{3}{c}{\textbf{Frame@30s}} & \multicolumn{3}{c}{\textbf{Frame@45s}} & \multicolumn{3}{c}{\textbf{Frame@59s}} \\
\cline{2-16}
 & \textbf{F1} & \textbf{P} & \textbf{R} & \textbf{F1} & \textbf{P} & \textbf{R} & \textbf{F1} & \textbf{P} & \textbf{R} & \textbf{F1} & \textbf{P} & \textbf{R} & \textbf{F1} & \textbf{P} & \textbf{R} \\
\hline
Inception-FCN & 15.48 & 60.00 & 8.89 & 15.67 & 51.09 & 9.25 & 13.92 & 43.43 & 8.29 & 11.56 & 41.98 & 6.71 & 7.75 & 39.29 & 4.30 \\
B-Inception-FCN & 17.74 & 62.50 & 10.34 & 19.84 & 58.72 & 11.94 & 18.71 & 51.69 & 11.42 & 17.84 & 55.34 & 10.63 & 10.63 & 51.67 & 5.93 \\
Xception-FCN & 87.25 & 86.82 & 87.69 & 87.27 & 85.28 & 89.36 & 86.58 & 84.14 & 89.16 & 87.10 & 84.82 & 89.50 & 75.65 & 74.12 & 77.25 \\
B-Xception-FCN & 91.43 & 89.72 & \textbf{93.20} & 90.15 & 86.80 & \textbf{93.78} & 90.12 & 87.69 & 92.70 & 90.35 & 87.64 & \textbf{93.22} & 88.30 & 84.24 & 92.77 \\
ResFCN & 88.73 & 89.71 & 87.77 & 89.43 & 89.76 & 89.10 & 90.43 & 91.38 & 89.50 & 88.81 & 88.69 & 88.92 & 87.10 & 90.23 & 84.17 \\
B-ResFCN & \textbf{93.29} & \textbf{95.16} & 91.50 & \textbf{92.55} & \textbf{91.52} & 93.61 & \textbf{93.62} & \textbf{94.02} & \textbf{93.22} & \textbf{93.06} & \textbf{94.33} & 91.83 & \textbf{94.54} & \textbf{95.28} & \textbf{93.81} \\
\bottomrule
\end{tabular}
\end{table*}

\subsection{Quantitative Evaluation}
To verify the effectiveness of networks used, we report in Table~\ref{tab:oa} pixel-level overall accuracies (OAs) and F1 scores \RR{}{of the car class} on our test set of ISPRS Potsdam dataset and compare to state-of-the-art methods. These metrics are calculated on a full reference and an alternative ground truth obtained by eroding the boundaries of objects by a circular disk of 3 pixel radius. \RR{}{The current state-of-the-art CASIA2 (in the leaderboard \url{http://www2.isprs.org/potsdam-2d-semantic-labeling.html}) obtains the F1 score of 96.2\% for the vehicle segmentation on the held-out test set (which is different from the validation set we use) using IRRG.} Our B-ResFCN is competitive with the F1 score of 95.87\% obtained by using RGB information only on our own test set. This indicates that the trained network can be though as a good, competitive model for the follow-up experiments. Note that the pixel-wise OA and F1 score can only evaluate the segmentation performance at \RR{}{pixel-level} instead of instance level. Therefore, they are actually not suitable for our task.
\par
To quantitatively evaluate the performance of different approaches for vehicle segmentation at instance level, the evaluation criteria we use are instance-level F1 score, precision, recall, and Dice similarity coefficient. The first three criteria consider the performance of vehicle detection, and the last validates the performance of instance-level segmentation.

\subsubsection{\textbf{Detection}}
For the vehicle detection evaluation, the metric instance-level F1 score\footnote{Note that the instance-level F1 score is different from the pixel-wise F1 score used by the ISPRS semantic labeling evaluation (\url{http://www2.isprs.org/commissions/comm3/wg4/semantic-labeling.html}).} is employed, which is the harmonic mean of instance-level precision P and recall R, defined as:
\begin{equation}\label{eq:f1}
F1=\frac{2PR}{P+R}\,,
P=\frac{N_{tp}}{N_{tp}+N_{fp}}\,,
R=\frac{N_{tp}}{N_{tp}+N_{fn}}\,,
\end{equation}
where $N_{tp}$, $N_{fp}$, and $N_{fn}$ are the number of true positives, false positives, and false negatives, respectively. Here, the ground truth for each segmented vehicle is the object in the manually labeled segmentation mask that has maximum overlap with the segmented vehicle. When calculating $N_{tp}$ and $N_{fp}$, a segmented vehicle that intersects with at least 50\% of its ground truth is considered a true positive; otherwise it is regarded as a false positive. For $N_{fn}$, a false negative indicates a ground truth object that has less than 50\% of its area overlapped by its corresponding segmented vehicle or has no corresponding segmented vehicle.
\par
The detection results of different networks on the ISPRS Potsdam dataset and Busy Parking Lot scene are shown in Table~\ref{tab:det_potsdam} and Table~\ref{tab:det_parkinglot}, respectively. Among the networks without semantic boundary component, ResFCN surpasses all other models (VGG-FCN, Inception-FCN, and Xception-FCN), highlighting the strength of residual learning-based FCN architecture with multi-level contextual feature representations in our task. The network with the semantic boundary component -- i.e., B-ResFCN -- achieved the best results on most test images of the ISPRS Potsdam scene and surpassed the others by a significant margin on the Busy Parking Lot dataset, demonstrating the effectiveness of the semantic boundary-aware multi-task learning network in this instance segmentation problem. From Table~\ref{tab:det_potsdam} and Table~\ref{tab:det_parkinglot}, we observe that all networks yield a fairly lower instance-level F1, precision, and recall on the Busy Parking Lot dataset than on the ISPRS Potsdam dataset. This mainly comes from the different difficulty levels of the two datasets. Specifically, high-density parking, strong light conditions, critical effects of shadow, and a slightly blurry image quality lead to the fact that networks achieved a more inferior performance on the proposed dataset than on the Potsdam scene.

\begin{table*}[t]
\caption{\label{tab:seg_parkinglot} Segmentation Results of Different Methods on Busy Parking Lot UAV Video Dataset (Instance-level Dice Similarity Coefficient)}
\centering
\begin{tabular}{cccccc}
\toprule
\textbf{Model} & \textbf{Frame@1s} & \textbf{Frame@15s} & \textbf{Frame@30s} & \textbf{Frame@45s} & \textbf{Frame@59s} \\
\hline
Inception-FCN & 26.81 & 26.06 & 25.68 & 22.89 & 23.77 \\
B-Inception-FCN & 32.37 & 33.07 & 33.34 & 30.44 & 31.26 \\
Xception-FCN & 72.74 & 72.74 & 72.85 & 72.47 & 71.31 \\
B-Xception-FCN & 77.31 & \textbf{77.50} & 77.22 & 77.13 & 76.32 \\
ResFCN & 71.17 & 71.47 & 71.76 & 68.82 & 72.73 \\
B-ResFCN & \textbf{78.84} & 77.33 & \textbf{79.13} & \textbf{77.83} & \textbf{79.39} \\
\bottomrule
\end{tabular}
\end{table*}

\begin{table}[t]
\caption{\label{tab:seg_potsdam} Segmentation Results of Different Methods on ISPRS Potsdam Semantic Labeling Dataset (Instance-level Dice Similarity Coefficient)}
\centering
\begin{tabular}{cccccc}
\toprule
\textbf{Model} & \textbf{2\_12} & \textbf{5\_12} & \textbf{7\_7} & \textbf{7\_8} & \textbf{7\_9} \\
\hline
VGG-FCN & 58.88 & 45.79 & 53.13 & 51.09 & 54.25 \\
B-VGG-FCN & 71.48 & 64.48 & 74.54 & 70.43 & 69.47 \\
Inception-FCN & 52.79 & 34.37 & 37.15 & 35.08 & 44.22 \\
B-Inception-FCN & 55.26 & 35.69 & 46.76 & 37.33 & 47.14 \\
Xception-FCN & 90.05 & 73.05 & 84.84 & 84.58 & 86.54 \\
B-Xception-FCN & 91.44 & 75.47 & 85.12 & 88.64 & 87.95 \\
ResFCN & 91.97 & 77.68 & 89.10 & 89.78 & 89.65 \\
B-ResFCN & \textbf{93.80} & \textbf{77.72} & \textbf{90.61} & \textbf{91.19} & \textbf{90.66} \\
\bottomrule
\end{tabular}
\end{table}

\subsubsection{\textbf{Segmentation}}
The dice similarity coefficient is often used to evaluate segmentation performance. Given a set of pixels $\bm{V}$ denoted as a segmented vehicle and a set of pixels $\bm{G}$ annotated as a ground truth object, the Dice similarity coefficient is defined as:
\begin{equation}\label{eq:dsc}
D(\bm{V},\bm{G})=\left. 2(|\bm{V}\cap\bm{G}|) \middle / (|\bm{V}|+|\bm{G}|) \right.\,.
\end{equation}
This, however, is not suitable for segmentation evaluation on individual objects (i.e., instance segmentation). Instead, in this paper, an instance-level Dice similarity coefficient is defined and employed as:
\begin{equation}\label{eq:ins_dsc}
D_{ins}(\bm{V},\bm{G})=\frac{1}{2}[\sum_{i=1}^{N_{V}}\omega_iD(\bm{V}_i,\bm{G}_i)+\sum_{j=1}^{N_{G}}\tilde{\omega}_jD(\tilde{\bm{V}}_j,\tilde{\bm{G}}_j)]\,,
\end{equation}
where $\bm{V}_i$, $\bm{G}_i$, $\tilde{\bm{G}}_j$, and $\tilde{\bm{V}}_j$ are the $i$-th segmented vehicle, ground truth object that maximally overlaps $\bm{V}_i$, $j$-th ground truth object, and segmented vehicle that maximally overlaps $\tilde{\bm{G}}_j$, respectively. $N_{V}$ and $N_{G}$ respectively denote the total number of segmented vehicles and ground truth objects. Furthermore, $\omega_i$ and $\tilde{\omega}_j$ are both coefficients and can be calculated as:
\begin{equation}\label{eq:ins_dsc_weight}
\omega_i=\frac{|\bm{V}_i|}{\sum_{k=1}^{N_V}|\bm{V}_k|}\,,
\tilde{\omega}_j=\frac{|\tilde{\bm{G}}_j|}{\sum_{k=1}^{N_G}|\tilde{\bm{G}}_k|}\,.
\end{equation}
\par
Table~\ref{tab:seg_potsdam} and Table~\ref{tab:seg_parkinglot} show the segmentation results of different approaches on the Potsdam scene and Busy Parking Lot dataset, respectively. We can see that our B-ResFCN achieves the best performance on both these two datasets. Compared to the ResFCN, there is a 1.16\% increment in terms of the instance-level Dice similarity coefficient on the Potsdam dataset and a 7.31\% improvement on the Busy Parking Lot scene. From the figures in these two tables, we can see that the networks offer a more inferior performance on the Busy Parking Lot dataset than on the Potsdam scene. This is also in line with our intention of proposing a more challenging benchmark dataset for the vehicle instance segmentation problem. In addition, it is \RR{}{worth noting} that basically all the networks with boundary components can offer better instance segmentations compared to those without boundary. This means that multi-task learning is useful for different CNN variants in our task.

\section{Conclusion}
\label{sec:conc}
In this paper, we propose a semantic boundary-aware unified multi-task learning residual fully convolutional network in order to handle a novel problem (i.e., vehicle instance segmentation). In particular, the proposed network harnesses multi-level contextual features learned from different residual blocks in a residual network architecture to produce better pixel-wise likelihood maps. We theoretically analyze the reason behind this. Furthermore, our network creates two separate, yet identical branches to simultaneously predict the semantic segmentation masks of vehicles and semantic boundaries. The joint learning of these two problems is beneficial for separating ``touching'' vehicles, which are often not correctly differentiated into instances. The network is validated using a large high-resolution aerial image dataset, ISPRS Potsdam Semantic Labeling dataset, and the proposed Busy Parking Lot UAV Video dataset. To quantitatively evaluate the performance of different approaches for the vehicle instance segmentation, we advocate using an instance-level F1 score, precision, recall, and Dice similarity coefficient as evaluation
criteria, instead of traditional pixel-wise overall accuracy (OA) and F1 score for semantic segmentation. Both visual and quantitative analysis of the experimental results demonstrate the effectiveness of our approach.

\section*{Acknowledgement}
The authors would like to thank the ISPRS for making the Potsdam data set available.

\ifCLASSOPTIONcaptionsoff
  \newpage
\fi

\bibliographystyle{IEEEtran}
\bibliography{reference}

\begin{IEEEbiography}[{\includegraphics[width=1in,height=1.25in,clip,keepaspectratio]{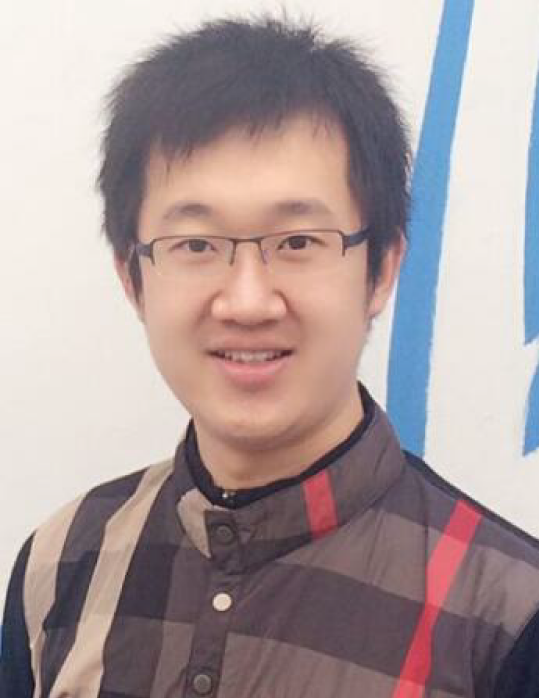}}]{Lichao Mou}(S'16)
received the bachelor's degree in automation from the Xi'an University of Posts and Telecommunications, Xi'an, China, in 2012, and the master's degree in signal and information processing from the University of Chinese Academy of Sciences, Beijing, China, in 2015. In 2015, he spent six months at the Computer Vision Group at the University of Freiburg in Germany. He is currently working toward the Ph.D. degree at the German Aerospace Center (DLR), Wessling, Germany, and the Technical University of Munich (TUM), Munich, Germany. His research interests include remote sensing, computer vision, and machine/deep learning, especially remote sensing video analysis and deep networks with their applications in remote sensing. He was the recipient of the first place in the 2016 IEEE GRSS Data Fusion Contest and a finalist for the Best Student Paper Award at the 2017 Joint Urban Remote Sensing Event.
\end{IEEEbiography}

\begin{IEEEbiography}[{\includegraphics[width=1in,height=1.25in,clip,keepaspectratio]{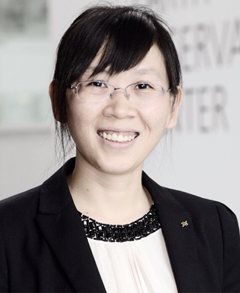}}]{Xiao Xiang Zhu}(S'10--M'12--SM'14) received the Master (M.Sc.) degree, her doctor of engineering (Dr.-Ing.) degree and her “Habilitation” in the field of signal processing from Technical University of Munich (TUM), Munich, Germany, in 2008, 2011 and 2013, respectively.
\par
  She is currently the Professor for Signal Processing in Earth Observation (www.sipeo.bgu.tum.de) at Technical University of Munich (TUM) and German Aerospace Center (DLR); the head of the department ``EO Data Science'' at DLR's Earth Observation Center; and the head of the Helmholtz Young Investigator Group ``SiPEO'' at DLR and TUM. Prof. Zhu was a guest scientist or visiting professor at the Italian National Research Council (CNR-IREA), Naples, Italy, Fudan University, Shanghai, China, the University  of Tokyo, Tokyo, Japan and University of California, Los Angeles, United States in 2009, 2014, 2015 and 2016, respectively. Her main research interests are
  remote sensing and Earth observation, signal processing, machine learning and data science, with a special application focus on global urban mapping.

  Dr. Zhu is a member of young academy (Junge Akademie/Junges Kolleg) at the Berlin-Brandenburg Academy of Sciences and Humanities and the German National  Academy of Sciences Leopoldina and the Bavarian Academy of Sciences and Humanities. She is an associate Editor of IEEE Transactions on Geoscience and Remote Sensing.
  \end{IEEEbiography}

\end{document}